\documentclass[conference]{IEEEtran}
\IEEEoverridecommandlockouts
\usepackage{cite}
\usepackage{amsmath,amssymb,amsfonts}
\usepackage{algorithmic}
\usepackage{graphicx}
\usepackage{textcomp}
\usepackage[acronym]{glossaries}
\usepackage{booktabs}
\usepackage{siunitx}
\usepackage{subfig}
\usepackage{mathtools}
\usepackage{tabularx,booktabs}
\usepackage[table]{xcolor}

\newcolumntype{C}{S<{\kern\tabcolsep}@{}}

\def\BibTeX{{\rm B\kern-.05em{\sc i\kern-.025em b}\kern-.08em
    T\kern-.1667em\lower.7ex\hbox{E}\kern-.125emX}}

\DeclarePairedDelimiterX\norm[1]\lVert\rVert{
\ifblank{#1}{\:\cdot\:}{#1}
}

\DeclarePairedDelimiter\abs\lvert\rvert
\reDeclarePairedDelimiterInnerWrapper\abs{star}{#1#2#3}
\reDeclarePairedDelimiterInnerWrapper\abs{nostarnonscaled}{\mathinner{#1#2#3}}
\reDeclarePairedDelimiterInnerWrapper\abs{nostarscaled}{\mathinner{#1#2#3}}

\providecommand\given{}
\newcommand\SetSymbol[1][]{%
\nonscript\:#1\vert
\allowbreak
\nonscript\:
\mathopen{}}
\DeclarePairedDelimiterX\Set[1]\{\}{%
\renewcommand\given{\SetSymbol[\delimsize]}
#1
}

\newacronym{cam}{CAM}{Class Activation Map}
\newacronym{coco}{COCO}{Common Object in COntext}
\newacronym{wsl}{WSL}{Weakly Supervised Learning}
\newacronym{cnn}{CNN}{Convolutional Neural Network}

\makeglossaries
\usepackage[bookmarks=false]{hyperref}

\hypersetup{
  breaklinks,
  colorlinks,
  pagebackref=true,
  backref=true,
  plainpages=false,
  pdfpagelabels,
  hypertexnames=false,
}
\begin{document}

\title{Weakly Supervised Semantic Segmentation of Satellite Images\\
}

\author{\IEEEauthorblockN{Adrien Nivaggioli}
\IEEEauthorblockA{{\itshape Qwant Research} \\
Paris, France \\
a.nivaggioli@qwant.com}
\and
\IEEEauthorblockN{Hicham Randrianarivo}
\IEEEauthorblockA{{\itshape Qwant Research}\\
 Paris, France\\
h.randrianarivo@qwant.com}
}

\maketitle

%        ---- IEEE COPYRIGHT for JURSE ----

\IEEEpubid{\begin{minipage}{\textwidth}\ \\[12pt] \centering
~ \\~ \\~\\
%-- CASE 1: For papers in which all authors are employed by the US government, uncomment the line below and comment CASE 4 line:
%U.S. Government work not protected by U.S. copyright % CASE 1?%-- CASE 2:   For papers in which all authors are employed by a Crown government (UK, Canada, and Australia), uncomment the line below and comment CASE 4 line:
% 978-1-7281-0009-8/19/\$31.00 \copyright 2019 Crown % CASE 2
%-- CASE 3:  For papers in which all authors are employed by the European Union, uncomment the line below and comment CASE 4 line:
% 978-1-7281-0009-8/19/\$31.00 \copyright 2019 European Union  % CASE 3
%-- CASE 4: For all other papers, default copyright:
978-1-7281-0009-8/19/\$31.00 \copyright 2019 IEEE  % CASE 4(default)
\end{minipage}}

\begin{abstract}

When one wants to train a neural network to perform semantic segmentation, creating pixel-level annotations for each of the images in the database is a tedious task.
If he works with aerial or satellite images, which are usually very large, it is even worse.
With that in mind, we investigate how to use image-level annotations in order to perform semantic segmentation.
Image-level annotations are much less expensive to acquire than pixel-level annotations, but we lose a lot of information for the training of the model.
From the annotations of the images, the model must find by itself how to classify the different regions of the image.
In this work, we use the method proposed by Anh and Kwak \cite{Ahn} to produce pixel-level annotation from image level annotation.

We compare the overall quality of our generated dataset with the original dataset.

In addition, we propose an adaptation of the AffinityNet that allows us to directly perform a semantic segmentation.

Our results show that the generated labels lead to the same performances for the training of several segmentation networks. Also, the quality of semantic segmentation performed directly by the AffinityNet and the Random Walk is close to the one of the best fully-supervised approaches.

\end{abstract}


\section{Introduction}
\label{sec:introduction}

Semantic segmentation of satellite and aerial images could be incredibly helpful for fields like urban planning, disaster recovery, autonomous agriculture, environmental monitoring and many others. We now have access to large databases filled with more images than any manual method could handle (such as the USGS Earth Explorer\footnote{\url{https://earthexplorer.usgs.gov/}}, ESA’s Sentinel Mission \footnote{\url{https://sentinel.esa.int/web/sentinel/home}} or NASA's Earthdata Search\footnote{\url{https://search.earthdata.nasa.gov/}}).
The need for an automatized study of those images is obvious, and
there is an urgent demand for tools and methods that allow automatic interpretation of this huge amount of data.
In the last few years, deep learning has become the essential tool for solving this kind of problem \cite{Krizhevsky2012,He2016}.
For the task of semantic segmentation, several successful methods appeared recently like SegNet \cite{Badrinarayanan2017}, FCN \cite{Shelhamer2016}, U-NET \cite{Ronneberger2015} or PSPNet \cite{Zhao2017a} with great results in satellite and aerial imagery \cite{Audebert2017a,Sherrah2016,Marcos2018}

One of the biggest difficulties faced by those methods is the lack of pixel-level labels for those images. Indeed, the process of manually annotating each image is tedious. This is also an issue when trying to perform a semantic segmentation of a regular image, like those of the Pascal VOC \cite{Everingham2014} or the \gls{coco} datasets \cite{Lin2014}. To face it, methods relying on weaker types of labels started to appear, called \gls{wsl}. Those methods can use simple bounding-box annotations \cite{Khoreva,  Papandreou2015}, scribbles \cite{Lin2016a, Vernaza2017a}, points \cite{Bearman2016}, or a simple image-level class label \cite{Oh2017,  Pinheiro2015}.

\begin{figure*}[htb]
\centering
\includegraphics[width=.9\textwidth]{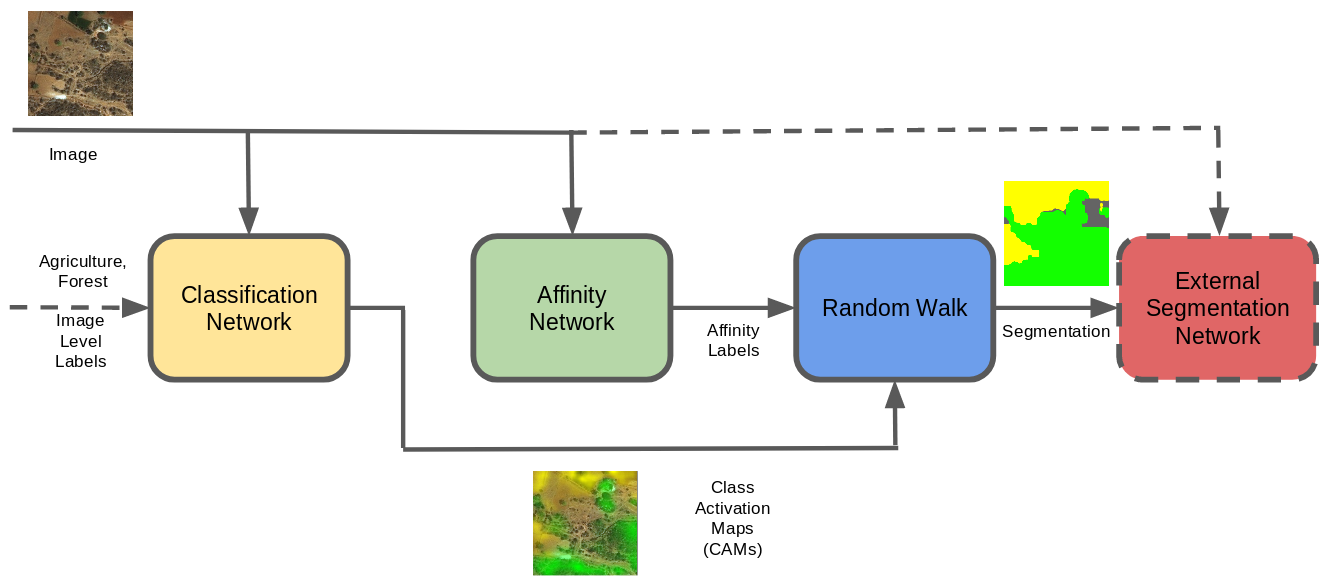}
\caption{Inference pipeline of the original Affinity-Net, that requires both an image and a label as inputs. We propose to remove the dotted parts to transform the Affinity-Net into an independent segmentation  network.}
\label{fig:overview}
\end{figure*}

The method we use is divided into 4 steps (c.f. fig.~\ref{fig:overview}), and based on the work of Ahn and Kwak \cite{Ahn}. First, a classification network is trained using image level annotations. Then, a second network is trained to learn the relationships between a pixel and its neighbourhood, called an affinity network. After this, a random walk is performed combining \gls{cam} and affinity labels to produce the segmentation labels. Finally, we can use the segmentation labels produced by the method to train a segmentation model.

{\bfseries Contributions:~}
We use a new backbone better adapted to satellite imagery.
We conduct several experiments on the loss function proposed by \cite{Ahn} to adapt its hyperparameters for satellite and aerial imagery.
We train several state-of-the-art segmentation models (SegNet \cite{Badrinarayanan2017}, PSPNet \cite{Zhao2017a}and UNet \cite{Ronneberger2015}) to validate the quality of the generated labels for the training of segmentation models.
We modify the method in order to perform semantic segmentation using only the classification and the affinity networks. (c.f. fig.~\ref{fig:overview})
Finally, we compare our semantic segmentation results with fully supervised approaches on the validation set of the DEEPGLOBE dataset \cite{Demir2018} and study the trade-off between information and quality of segmentation..

\section{Method}
\label{sec:method}

In this section, we present the approach we follow for weakly based semantic segmentation. The method can be split into 4 different parts as shown in fig.~\ref{fig:overview}~: Classification Network, Affinity Network, Random Walk and Segmentation Network.

\subsection{Classification Network}

The first step of the method is the training of a classification network that will be used to identify the different categories present in the images. The main idea is to use the capacity of the \gls{cam} that can be extracted from the \gls{cnn} to localize in the image the areas that influence the most the classification result \cite{Pinheiro2015,Zhou2016,Bazzani2016}.  For each image, the classification network will produce a set of \glspl{cam} $\mathcal{M} = \Set{M_c \given c \in \mathcal C}$, each \gls{cam} $M_c$ corresponding to the activation map of the class $c$. Additionally, another \gls{cam} $M_{bg}$ is defined to localize background in the image~\eqref{eq:cam_bg}.

\begin{equation}
\label{eq:cam_bg}
 M_{bg}(x,y) = \Set{ 1 - \max_{c \in \mathcal{C}} M_c(x,y) }^\alpha
\end{equation}

Usually, there is no background in aerial images, but this \gls{cam} prevents the network to predict unsure result. Even if this means that a smaller portion of the original dataset will be usable to train the segmentation network, we favour precision over quantity.  With that in mind, we conducted all of our experiments both with and without the background, in order to compare the results and assess which approach fits our case best.

\subsection{Affinity Network}

The second part of the method consists in training an Affinity Network which models relationships between pairs of pixels $(i, j)$ in the image. The AffinityNet is designed to extract a convolutional feature map $f^{\text{aff}}$ where each element can be seen as an affinity feature. The affinity between two pixels $(i, j)$ denoted by $W_{ij}$ is defined as the similarity between the affinity features:

\begin{equation}
    W_{ij} = \exp\Set{ -\norm{f^{\text{aff}}(x_i, y_i) - f^{\text{aff}}(x_j, y_j)}_1 }
\end{equation}

\begin{figure}[htb]
\centering
\includegraphics[width=.75\linewidth]{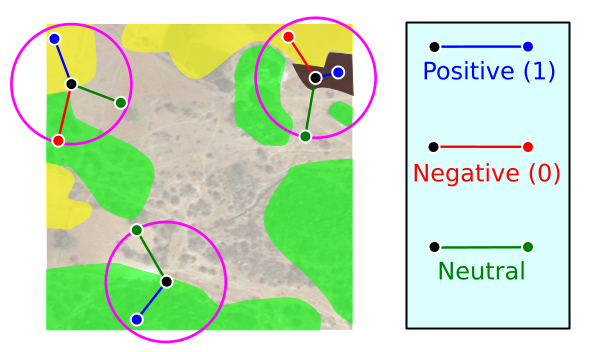}
\caption{Generating Affinity labels. The areas correspond to confident prediction of \textit{Agriculture} in yellow and  \textit{Forest} in green. The black area is the \textit{background} and the white is \textit{neutral}.}
\label{fig:affinity}
\end{figure}

Nevertheless, the affinity labels needed to train the AffinityNet are not directly available.
A way to generate these affinity labels is to use the \gls{cam} as partial sources of supervision for their generation. The values of the \gls{cam} are used as a confidence score to determine whether a pixel belongs to a category $c$ or not. A high $alpha$ is used to assess confident regions of classes $c \in \mathcal{C}$, and a lower one is used to assess areas the network is the most unsure about . The pixels that are left are considered as neutral. Using the \gls{cam} with high confidence, a binary label is given to the pair of pixels. If theirs classes are the same, their affinity label $W^*_{ij}$ is $1$, $0$ otherwise. The pairs containing neutral pixel are considered neutral and ignored during the training (c.f. fig.~\ref{fig:affinity}).

From there, we compute for each coordinate of the image its affinities with the other coordinates within a circle of fixed radius $\gamma$. The set of pairs of coordinates that have an affinity is defined by~:

\begin{equation}
\label{eq:affinity_labels}
\mathcal{P} = \Set{ (i,j) \given  d((x_i,y_i),(x_j,y_j)) <  \gamma, \forall i \ne j }
\end{equation}

The set $\mathcal{P}$ is then divided into two subset $\mathcal{P}^+$ and $\mathcal{P}^-$ where $W^*_{ij} = 1$ and $W^*_{ij} = 0$ respectively. The subset $\mathcal P^+$ is again divided into two subsets $\mathcal P^+_{fg}$ for foreground and $\mathcal P^+_{bg}$for the background regions.

The model is trained using the following loss function~:
\begin{align} \label{eq:loss}
\mathcal{L} &= \frac{\mathcal{L}^+_{fg}}{a} + \frac{\mathcal{L}^+_{bg}}{b} +  \frac{\mathcal{L^-}}{c}\\ \intertext{with}
\dfrac{1}{a} &+\dfrac{1}{b} +\dfrac{1}{c} = 1\label{eq:parameters}
\end{align}

\begin{align}
\mathcal{L}^+_{fg}  &= - \frac{1}{\abs{ \mathcal{P}_{fg}^+ } } \sum_{(i,j) \in \mathcal{P}_{fg}^+} \log{W_{i,j}} \\
\mathcal{L}^+_{bg}  &= - \frac{1}{\abs{ \mathcal{P}_{bg}^+ } } \sum_{(i,j) \in \mathcal{P}_{bg}^+} \log{W_{i,j}} \\
\mathcal{L}^-  &= - \frac{1}{\abs{ \mathcal{P}^- } } \sum_{(i,j) \in \mathcal{P}^-} \log{(1- W_{i,j})}
\end{align}

This loss is quite similar to the one used by the authors of\cite{Ahn}, but they suggest that $a$ should be equal to $b$. We want to be able to penalize the background further, so we slackened the constraints. \cite{Ahn} used eq.~\ref{eq:loss} with $a=b=4, c=2$. In order to penalize the background further, we used $a=6, b=2, c=3$.

Those labels can then be used to perform a random walk on the original \glspl{cam} in order to get proper segmentation labels.

From there, the original method uses those labels to train a regular segmentation network. We compared the results of 3 different networks, U-Net \cite{Ronneberger2015}, PSPNet \cite{Zhao2017a} and SegNet \cite{Badrinarayanan2017} using our generated labels.

We also tried a new approach that allows us to directly performs semantic segmentation with the trained classification and affinity networks. Indeed, \cite{Ahn} proposes to set to 0 the \glspl{cam} of the class we know are not present in the image. While this increases the quality of the segmentation labels, it prevents us from using the AffinityNet on images we do not have any labels on. To assess which \gls{cam} need to be kept, we use the confidence scores output by the classification network.

As a backbone network, we used an extension of the ResNet38 \cite{Wu2016}, composed of 74 convolutional layers (as opposed to 38 in the ResNet38). We found that a deeper network greatly improved our performances, supposedly because aerial images have a lower variance than regular ones, thus making it more difficult for the  network to differentiate the different classes.

\section{Experiments and Results}
\label{sec:experiments}

\subsection{Dataset}
We used the DeepGlobe dataset \cite{Demir2018} composed of 803 aerial images with pixel-level labels. We kept 562 for training and the remaining 241 for validation. Each image has a size of 2448x2448 pixels, and contains at least one of those categories : Barren, Water, Urban, Forest, Agriculture, Rangeland and Unknown. We split each image into 64 patches of size 306x306 each. Even if the dataset comes with pixel-wise annotations of the images, we reduced those labels to image-level (which categories are present in the labels). This means that no information about localization and/or distribution of the categories are present in the final dataset. The DeepGlobe dataset also offers 171 aerial images without any labels that we used for testing. Indeed, we can upload our results to their CodaLab competition\footnote{\url{https://competitions.codalab.org/competitions/18468}} and compare our score to others, all of them using fully-supervised techniques.

\subsection{Quality of Segmentation Labels}
\begin{figure}[htpb]

  \centering

  \subfloat{\includegraphics[width=.24\linewidth]{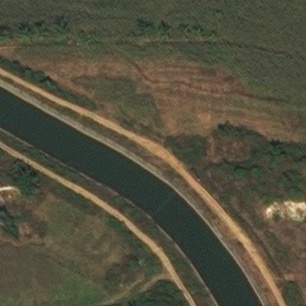}}
  \hfill
  \subfloat{\includegraphics[width=.24\linewidth]{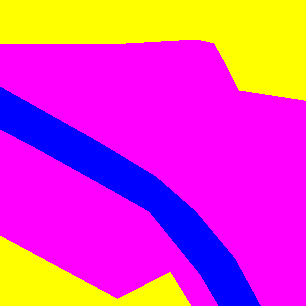}}
  \hfill
  \subfloat{\includegraphics[width=.24\linewidth]{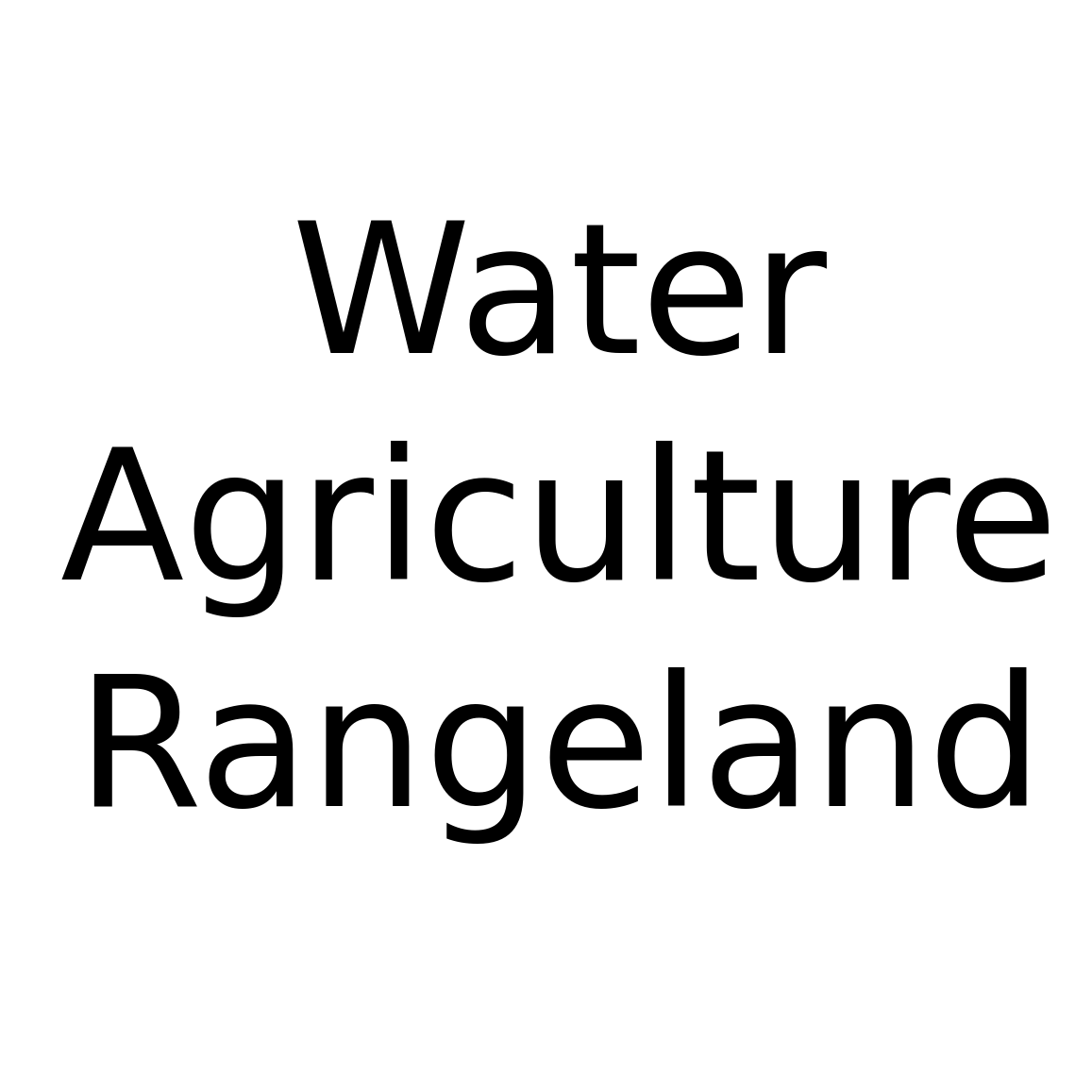}}
  \hfill
  \subfloat{\includegraphics[width=.24\linewidth]{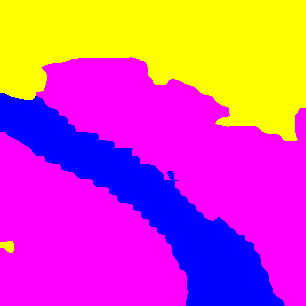}}
  \\[-1ex]
  \subfloat{\includegraphics[width=.24\linewidth]{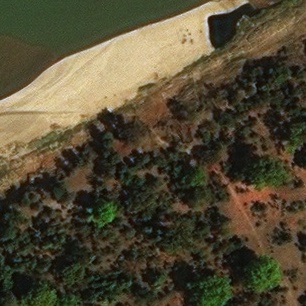}}
  \hfill
  \subfloat{\includegraphics[width=.24\linewidth]{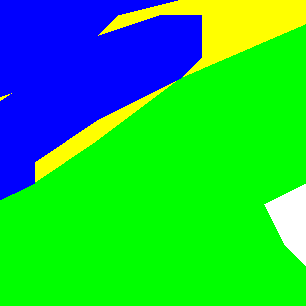}}
  \hfill
  \subfloat{\includegraphics[width=.24\linewidth]{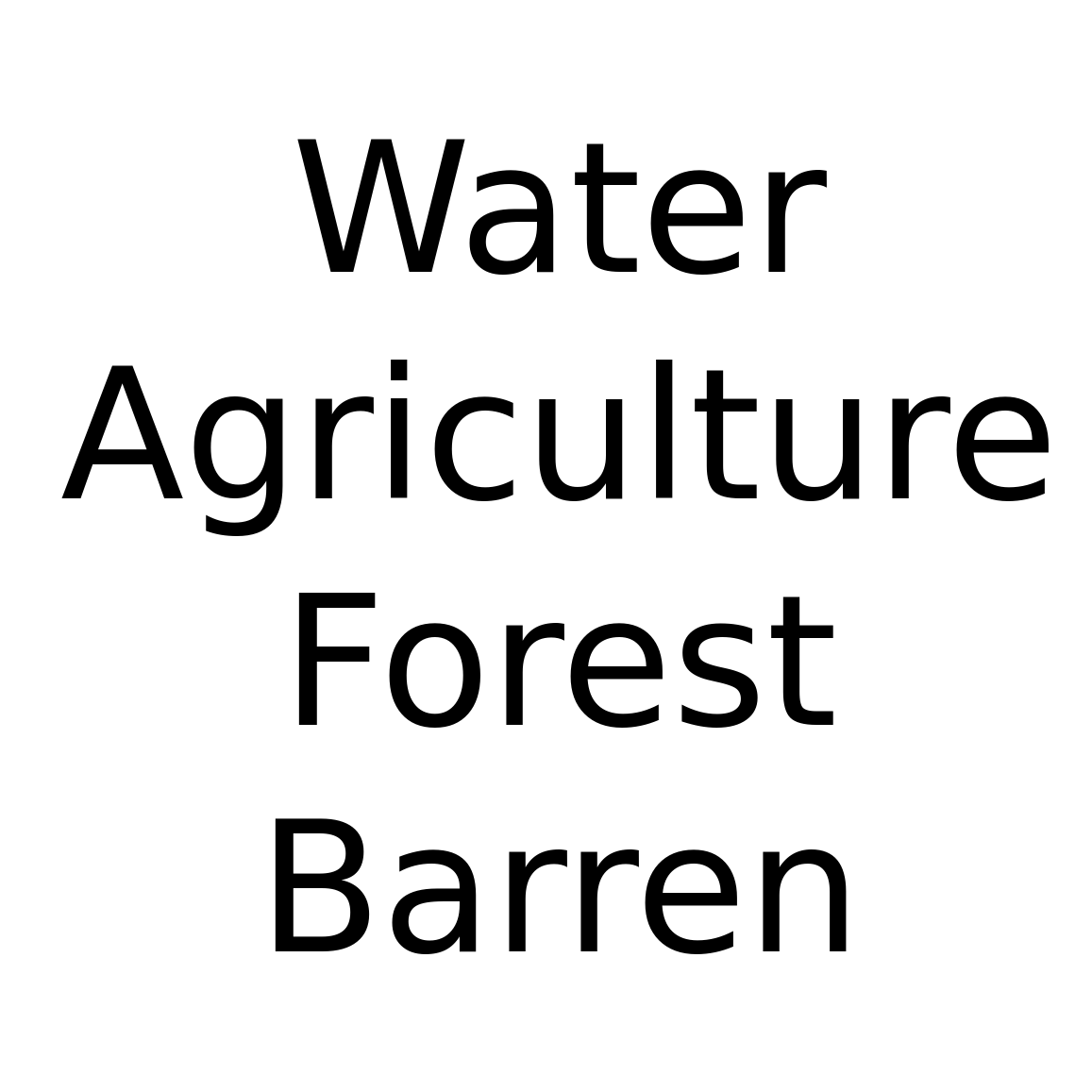}}
  \hfill
  \subfloat{\includegraphics[width=.24\linewidth]{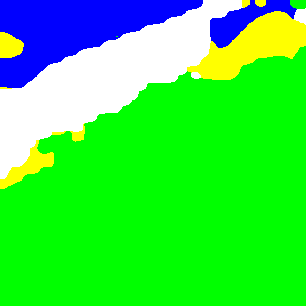}}
  \\[-1ex]
  \subfloat{\includegraphics[width=.24\linewidth]{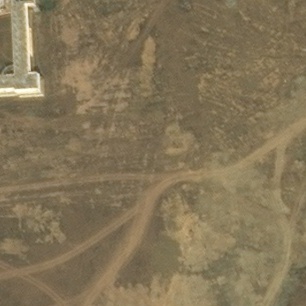}}
  \hfill
  \subfloat{\includegraphics[width=.24\linewidth]{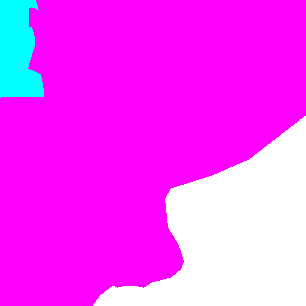}}
  \hfill
  \subfloat{\includegraphics[width=.24\linewidth]{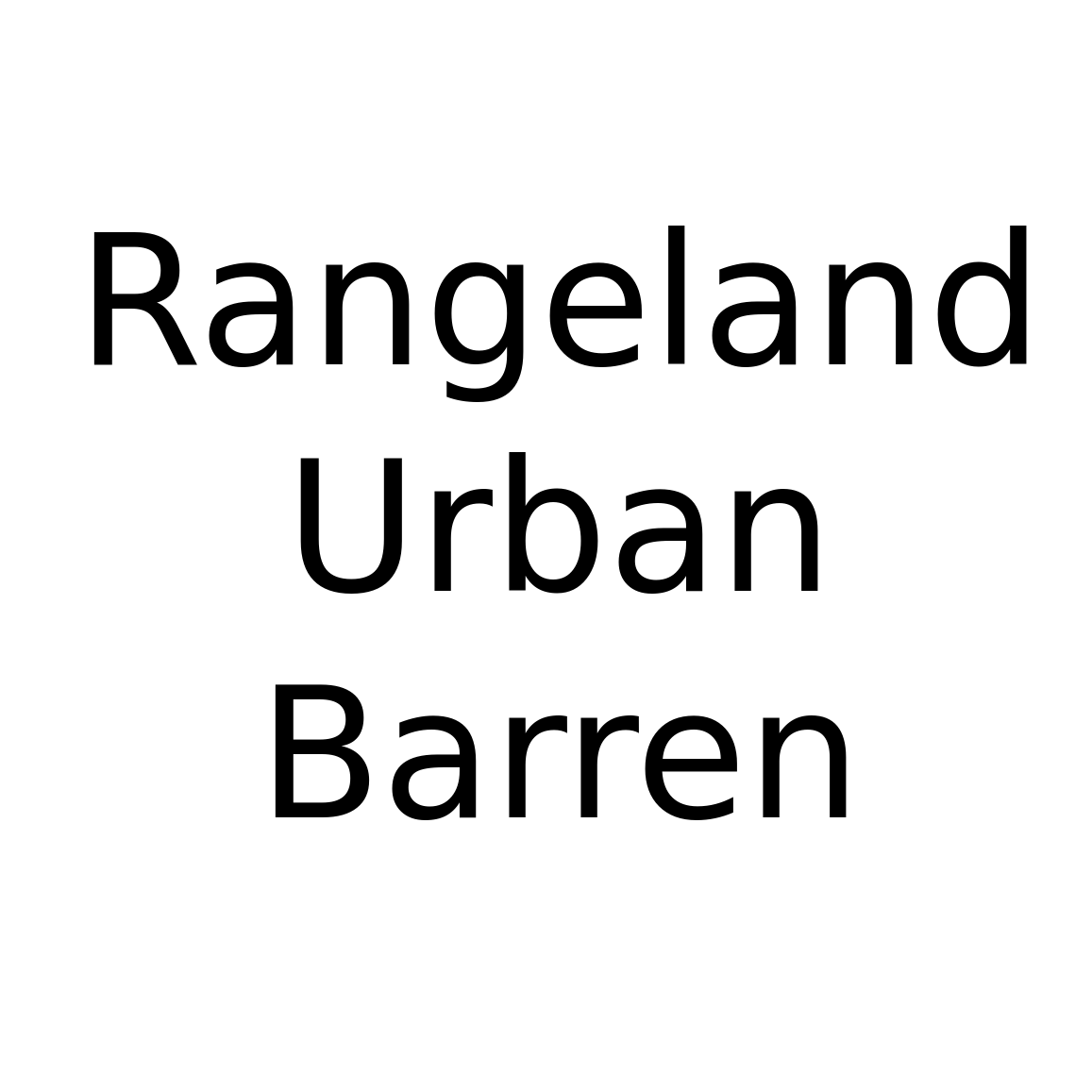}}
  \hfill
  \subfloat{\includegraphics[width=.24\linewidth]{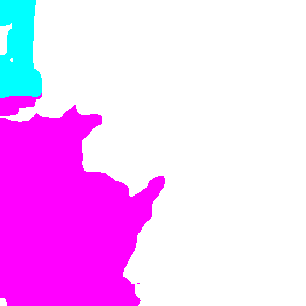}}
  \caption{Example of segmentation labels. From left to right: Original Image, Ground Truth Label, Our Image-Level label,  Predicted Label. Note that in order to generate the predicted label, we used only the image-level annotations and original images.\label{fig:seglabels}}
\end{figure}

We used the training dataset to teach both the classification Network and the Affinity Network. Then, we generated the Segmentation Label from the validation dataset. Because we have the pixel-level labels of those images (even if they were not used during the training), we can compare the quality of our segmentation labels to the ground truth provided by DeepGlobe as shown in fig.~\ref{fig:seglabels}. \cite{Ahn} showed that their method was fairly insensitive to hyperparameters, but our dataset of aerial images is different from regular images, mostly because they have no background class. With that in mind, we tried modifying the loss of the Affinity-Net, decreasing the background \glspl{cam}, and removing the background completely. We measured the precision and the recall of the predictions. The results showed that removing the background altogether gives almost the best precision, but with a far better recall (cf Table. \ref{table:labels}).

\begin{table}[htbp]
\centering
\caption{Quality of Segmentation Labels produced by the AffinityNet. Parameters of eq.~\ref{eq:parameters} are $a=6$, $b=2$, $c=3$. $\alpha=\infty$ means background has not be taken into account.}
\rowcolors{2}{gray!15}{white}
\begin{tabular}{SSS}
\toprule
$\alpha$ & {Precision}  & {Recall}\\
\midrule
 4 & 85.22 &  60.40     \\
 16 & 86.52 &  66.47     \\
 32 & 86.85 & 67.28 \\
$\infty$ & 88.38 & 87.31 \\
\bottomrule
\end{tabular}
\label{table:labels}
\end{table}

Furthermore, we use the best generated labels to train 3 different segmentation networks, and compare the results with a fully-supervised training on the same networks (cf Table. \ref{training}).

\begin{table}[htbp]
\caption{Results of semantic segmentation on the DEEPGLOBE land classification challenge. Full labels are labels provided for the challenge . Weak labels are labels produced using weakly supervised model.}
\centering
\rowcolors{2}{gray!15}{white}
\begin{tabular}{ccS}
\toprule
\multicolumn{2}{c}{Method} & \text{Score (mIoU)} \% \\
\midrule
PSPNet & Full labels &   43.99    \\
- & Weak labels &   42.97    \\
U-Net & Full labels  &  42.44    \\
- & Weak labels  &  39.25    \\
SegNet & Full labels &  37.10    \\
- & Weak labels  &    37.76 \\
\bottomrule
\end{tabular}
\label{training}
\end{table}

\subsection{Semantic Segmentation}
If we follow the method described in \cite{Ahn}, we cannot create the Segmentation Labels without providing the image-level labels, even after training. We modified the method of \cite{Ahn} to directly perform the segmentation, without having to train a separate segmentation network.
We were able to evaluate our results and compare them to to the best, fully-supervised approaches of the deepglobe competition (cf Table. \ref{tab2}).

\begin{table}[htbp]
\caption{Ranking of our Weakly-Supervised method among Fully-Supervised ones}
\centering
\rowcolors{2}{gray!15}{white}
\begin{tabular}{cSS}
\toprule
\text{Method} & \text{Score (mIoU)} & \text{Rank}\\
\midrule
Deep Aggregation Net  &  53.58 & 1      \\
Dense Fusion Classmate Network & 52.64 & 2 \\
Ours - Weakly Supervised w/o background & 45.90 & 14\\
Ours - Weakly Supervised w/ background  &32.32 & 33\\

\bottomrule
\end{tabular}
\label{tab2}
\end{table}

\section{Conclusion}
\label{sec:conclusion}

We adapted the method proposed in \cite{Ahn} to our dataset composed of aerial images. The quality of the pixel-level labels generated from  image-level labels are quite good. Both offer similar performances when used for training several segmentation network. This means that less information in a dataset does not necessarily implies inferior results.

Furthermore, the semantic segmentation performed by our modification of the AffinityNet gives remarkable results, close to best fully supervised ones.

Further work will be done to improve our results, both on the quality of the generated segmentation labels and the direct semantic segmentation.

Weakly supervised semantic segmentation of satellite images is a cornerstone to the transition from geolocalized text information to complete semantic segmentation.

\bibliographystyle{plain}
\bibliography{Jurse2019}

\end{document}